\documentclass[letterpaper, 10 pt, journal, twoside]{IEEEtran}

\usepackage{graphics} 
\usepackage{graphicx}
\usepackage{epsfig} 
\usepackage{mathptmx} 
\usepackage{times} 
\usepackage{amsmath} 
\usepackage{amssymb}  
\usepackage{rotating}
\usepackage{array}
\usepackage{multirow}
\usepackage{makecell}
\usepackage{multicol}
\usepackage{booktabs}
\usepackage{makecell}
\usepackage{colortbl}
\usepackage{hyperref}
\usepackage{algorithm}
\usepackage{algorithmic}
\usepackage{marvosym}

\usepackage{geometry}
\begin{document}
%
\title{GMM: Delving into Gradient Aware and Model Perceive Depth Mining for Monocular 3D Detection}
\author{Weixin Mao$^{*1}$, Jinrong Yang$^{*2}$, Zheng Ge$^{1}$, Lin Song${^3}$, \\
Hongyu Zhou$^{4}$, Tiezheng Mao${^1}$, Zeming Li$^{3}$, Osamu Yoshie$^{1}$%
\thanks{\emph{($^{*}$These authors contributed equally. Corresponding author: Osamu Yoshie.)}

$^{1}$Weixin Mao, Zheng Ge, Tiezheng Mao, Osamu Yoshie are with the Waseda University, Tokyo, 169-8050, Japan. {\tt\footnotesize (e-mail: maowx2017@fuji.waseda.jp, jokerzz@fuji.waseda.jp, maotiezheng@asagi.waseda.jp, yoshie@waseda.jp)}

$^{2}$Jinrong Yang is with the Huazhong University of Science and Technology, Wuhan, 430074, China. {\tt\footnotesize (e-mail: yangjinrong@hust.edu.cn)}

$^{3}$Lin Song is with the Xi’an Jiaotong University, Xi'an, 710049, China. {\tt\footnotesize (e-mail: stevengrove@stu.xjtu.edu.cn)}

$^{4}$Hongyu Zhou,  Zeming Li are with the MEGVII Technology, Beijing, 100096, China. {\tt\footnotesize (e-mail: zhouhongyu@megvii.com, lizeming@megvii.com)}

Digital Object Identifier (DOI): see top of this page.
}
}

\markboth{IEEE Robotics and Automation Letters. Preprint Version. May 2023}
{MAO \MakeLowercase{\textit{et al.}}: GMM: Delving into Gradient Aware and Model
Perceive Depth Mining for Monocular 3D Detection}

\maketitle

\begin{abstract}

Depth perception is a crucial component of monocular 3D detection tasks that typically involve ill-posed problems. In light of the success of sample mining techniques in 2D object detection, we propose a simple yet effective mining strategy for improving depth perception in 3D object detection. Concretely, we introduce a plain metric to evaluate the quality of depth predictions, which chooses the mined sample for the model. Moreover, we propose a Gradient-aware and Model-perceive Mining strategy (GMM) for depth learning, which exploits the predicted depth quality for better depth learning through easy mining. GMM is a general strategy that can be readily applied to several state-of-the-art monocular 3D detectors, improving the accuracy of depth prediction. Extensive experiments on the nuScenes dataset demonstrate that the proposed methods significantly improve the performance of 3D object detection while outperforming other state-of-the-art sample mining techniques by a considerable margin. On the nuScenes benchmark, GMM achieved the state-of-the-art (\textbf{42.1\%} mAP and \textbf{47.3\%} NDS) performance in monocular object detection.

\end{abstract}


\IEEEpeerreviewmaketitle

\section{Introduction} \label{Introduction}

\IEEEPARstart{3}{D} object detection has made significant progress in recent years, focusing on identifying objects with 3D size, location, pose, and category. Perception sensors such as LiDAR or cameras are often used to detect 3D objects and are applied in various domains, including autonomous driving and robotics. Due to their simplicity and cost-effectiveness, monocular 3D object detection methods have become a hot research topic. Most mainstream monocular detectors~\cite{fcos3d, pgd} follow the dense prediction framework of 2D object detectors~\cite{fcos, retinanet} and scale to the 3D prediction by appending additional task head (i.e., depth, rotation, attribution, etc.). Although these methods have achieved remarkable progress and decent performance, a notorious challenge remains in forecasting depth in images, making them sub-optimal.


Perceiving depth is an \emph{ill-posed issue} in monocular prediction, as noted by previous studies such as ~\cite{fcos3d, pgd, monoDis}. To address this issue, FCOS3D~\cite{fcos3d} uses a naive Smooth L1 loss to train the depth branch, while PGD~\cite{pgd} emphasizes that directly estimating depth based on isolated instances or pixels is sub-optimal since it ignores the geometric relations among different objects. As a solution, PGD proposes employing geometric relation graphs across predicted objects along with a probabilistic representation to capture depth uncertainty. However, these methods mainly focus on the depth representation paradigm and rarely explore how to \emph{optimize depth learning}.

Building on the aforementioned gap, we primarily focus on exploring an effective and orthogonal depth optimization algorithm. Specially, we notice that existing monocular 3D detector employs a  \emph{dense prediction} framework, i.e., they set a mass of anchor (points) to act as prediction samples. This approach presents a challenge since the biggest issue with a dense prediction framework is determining how to handle all samples differently. This approach presents a challenge since the biggest issue with a dense prediction framework is determining how to handle all samples differently. Another meaningful way is to carry out the sample-mining strategy. In this regard, depth prediction has not been explored yet. We first compare existing solutions from the perspective of sample mining for depth. They primarily include easy~\cite{PISA,iou-balanced} and hard~\cite{streamyolo1, streamyolo2} mining manners, which aim to re-weight different samples with subjective quality metrics (e.g., artificially defined location accuracy) for optimal learning. Thus, we can also refer to it as Subjective Mining. As shown in Table~\ref{table: fcos3d_reweight}, all mining schemes show subtle improvements but not enough outstanding. To this end, it poses a promising challenge: \emph{designing a more effective sample mining strategy for depth learning}.

To address this question, we first expose the potential shortcomings of the traditional sample mining methods mentioned earlier. These methods actually \emph{intensify} the degree of the depth ill-posed problem. They assign more weight to samples with subjectively high depth quality, but these samples may be ill-posed and should be treated as outliers. Focusing more on outlier samples can lead to sub-optimal learning, which is why the performance gains from traditional methods are limited.


In this paper, we tackle this challenge from two dimensions: model self-perception depth mining and gradient-aware depth mining. Our focus is on transforming the depth prediction into a \emph{soft} ill-posed problem, with the primary objective of capturing outliers. To achieve this, we introduce Model self Perception depth Mining (MPM), which appends a task head to predict the quality of depth. Specially, we design a depth quality factor to evaluate the accuracy of prediction depth, with its value constrained between [0, 1]. Subsequently, we utilize the model's own perceived depth quality to carry out easy mining. \textbf{This approach effectively eliminates the influence of outliers, as poorly predicted samples are the actual outliers.} Furthermore, we introduce Gradient-Aware depth Mining (GAM) to incorporate the sample mining procedure into a single loss function, which only needs to use the simple binary cross entropy (BCE) loss to achieve the above two motivation effect. Moreover, the predicted depth quality can be adapted to integrate with naive metrics (e.g., classification and centerness) of NMS, making it reasonable to pay more attention to depth in the NMS procedure.

We conduct extensive experiments on the nuScenes dataset~\cite{nuscenes}, showing the effectiveness of our method. In summary,  the contributions of this work are as three-fold as follows:
\begin{itemize}
\item We conduct extensive experiments to reveal that the depth sample mining strategy can help depth learning but show limited improvement. We traced the problem to a culprit: a subjective mining strategy will pay more attention to outliers.

\item We introduce the MPM module to weaken the influence of outliers and give full play to the power of the sample mining strategy. Furthermore, GAM is designed to collectively carry out outliers perception and sample mining procedures. Moreover, the prediction depth quality can be embedded into the NMS process to boost performance further.

\item Our proposed GMM achieves monocular 3D object detection SOTA performance on the nuScenes dataset, which outperforms other sample mining methods by a large margin. Especially, our sample mining way is orthogonal to existing depth supervision methods. 
\end{itemize}

\section{RELATED WORK}
\subsection{Camera-based 3D Object Detection}
In recent years, 3D object detection has made significant breakthroughs with autonomous driving and robotics advancements. Depending on the sensor, 3D object detection can be broadly classified as 3D point cloud object detection, monocular 3D object detection, or multi-camera 3D object detection, whereas monocular and multi-camera detection can be classified as image-based 3D target detection. In general, image-based 3d object detectors use RGB images as input and combine intrinsics and extrinsic of camera parameters for 3D bounding box prediction.

As a widely used 3D detection method, monocular 3D object detection has recently emerged with many excellent solutions. MonoDIS~\cite{monoDis} proposes decoupled regression losses to improve multi-task training. Imitating the work of the anchor-free 2D object detection FCOS~\cite{fcos}, ~\cite{fcos3d} proposes FCOS3D, which projects the 3D ground truths into image view to predict depth, sizes, offsets, yaw angles and then projects back into 3D space to obtain 3D bounding boxes. PGD~\cite{pgd} is based on FCOS3D and employs geometric restrictions and depth probabilities to increase depth estimate accuracy. It considerably relieves the depth estimation problem while increasing the compute budget and increasing inference delay. In addition, there are quite a bit of work projecting images into the bird's-eye-view (BEV) space to do 3D object detection, such as OFT~\cite{OFT}, CADDN~\cite{CaDDN}. DD3D~\cite{DD3D} demonstrates that depth pre-training on large-scale depth datasets can greatly improve 3D object detection performance. Some work has used DD3D's pre-trained model as a backbone to improve model performance.

Recently, work with multiple cameras is also very promising. Lif-Splat-Shoot(LSS)~\cite{lss} using monocular depth estimation to project 2D picture features into a per-camera frustum of view and splat them on BEV. After this, a series of work based on LSS, such as~\cite{bevdet, bevdet4d, bevfusion, bevseg, fiery, m2bev, li2022bevdepth}, etc. 


\subsection{Camera Depth Estimation}
An essential difference between 2D and 3D object detection is that 2D object detection does not require depth estimation and an accurate depth often determines the upper limit of the model. Hence, ~\cite{monoflex, pgd, CaDDN, DORN, GUPNet} devote a great deal of attention to improving depth estimates. MonoFlex~\cite{monoflex} calculates each of the three sets of 2D heights by introducing ten keypoints to obtain each of the three sets of depths. The PGD~\cite{pgd} model builds a depth propagation map between these detection targets with uncertainty through perspective geometric relationships, enhancing the accuracy of depth estimation with global information. CaDDN~\cite{CaDDN} considers the difficulty of predicting the depth of continuous values and tries to discretize the depth by converting it into a classification problem with discrete depth values. DORN, CaDDN turns depth estimation into a classification problem, where the number of categories is determined by dividing the farthest practical distance into multiple bins.

\begin{figure*}[t]
\centering
\includegraphics[width=\linewidth]{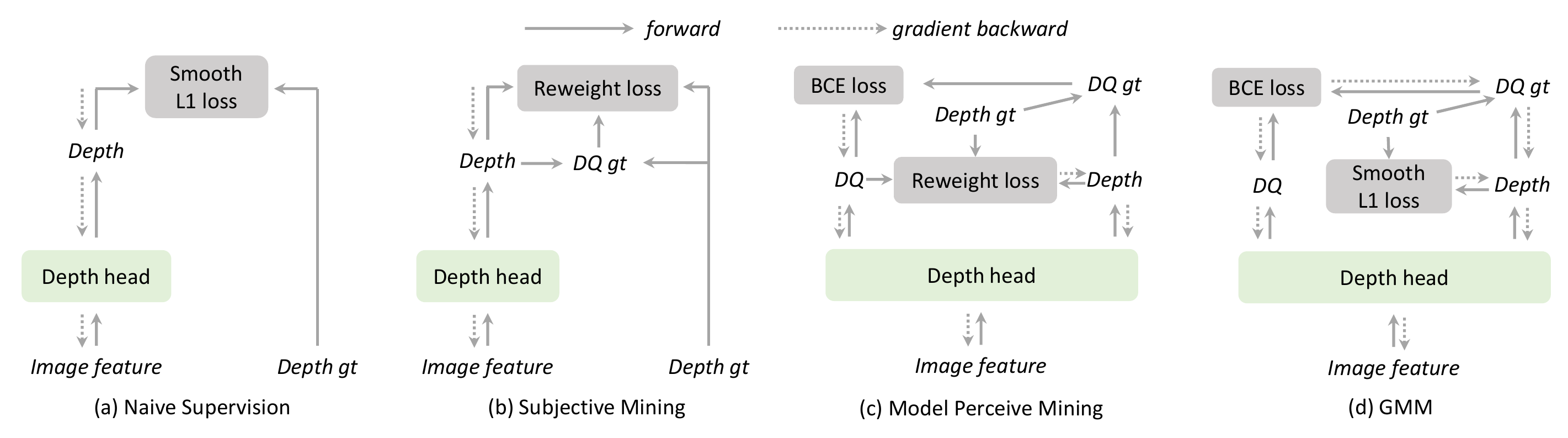}
\vspace{-.2in}
\caption{\textbf{Conceptual comparison of four depth supervision frameworks.} (a) The depth prediction is supervised by ground truth depth using naive smooth L1 loss. (b) Depth loss is re-weighted by the depth quality calculated between prediction depth and ground truth depth, which is a subjective mining way. The DQ ground truth consists of depth predictions and depth ground truths. (c) \textbf{MPM (Ours):} the depth head extra forecasts the depth quality for re-weighting the depth loss, which belongs to model self-perceive mining manner. (d) \textbf{GMM (Ours):} naive depth loss is intact while BCE loss is designed to supervise depth quality prediction and execute model self-perceive mining.}
\label{fig:framework}
\end{figure*}

\subsection{Sample Mining Strategy}

Sample mining strategy is a crucial techniques in the detection task, especially the dense prediction framework. It focuses on applying distinguishing supervisions to different prediction samples by virtue of various transcendental information. Several efforts advocate for easy sample mining. They introduce a learning framework to guide the model to focus more on samples with better accuracy. PISA~\cite{PISA} and IoU-balanced RetinaNet~\cite{iou-balanced}, as this faction, adopt IoU to measure the prediction quality of boxes and structure reweight loss for training detection models. Some works start with gradient analysis, aiming to balance gradient distribution for training samples. Libra R-CNN~\cite{libra_r-cnn} analyses that outliers occupy most of the gradient, so it designs a balanced loss to assign more gradients for easy samples. Like this viewpoint, GHM~\cite{GHM} pursues absolute gradient equilibrium. It first calculates the gradient density distribution and then modulates specifically one balanced loss for sample training. In addition, focal loss~\cite{retinanet} is proposed to tackle the imbalance between hard and easy samples, which also alleviates the imbalance between positive and negative samples. For other detection tasks, the sample mining strategy is also versatile. StreamYOLO~\cite{streamyolo1,streamyolo2} aims to tackle the streaming perception issue in a dense prediction framework, which employs a hard mining way to guide the model to pay attention to faster-moving objects. In short, the sample mining strategy is effective in 2D tasks, but it is rarely researched in 3D detection tasks. In this paper, we explore using this techniques endows a 3D monocular detector with optimal depth learning or perception. Specially, we introduce a whole new perspective on model-aware mining ways to transfer ill-posed issues to soft ill-posed ones for depth learning.


\subsection{Learning Prediction Quality} 

In the field of 2D/3D object detection, the notorious imbalance issue between NMS metrics (e.g., classification score, centerness or IoU) and true prediction accuracy is a headache. Therefore, some researchers propose to predict the prediction accuracy again to participate in the NMS procedure. IoU-Net~\cite{iounet} and IoU-aware~\cite{iou-aware} append a parallel branch to predict IoU between prediction boxes and corresponding target boxes, then adopt the prediction IoU to incorporate with classification score for the NMS process.~\cite{scd} indirectly tackle this issue by predicting the logit gap between classification score and location accuracy. FCOS~\cite{fcos} and FCOS3D~\cite{fcos3d} structure a variant, i.e., centerness score, to alleviate the imbalance problem. For instance, the segmentation task, Mask Scoring R-CNN~\cite{mask} employs the same mechanism to tackle this imbalance problem in mask prediction. To avoid adding an extra branch,~\cite{varifocalnet,gfl} directly predicts a unified score that fuses classification and location quality scores for the NMS process. In this paper, we focus on depth learning and propose to design a metric to measure depth prediction quality. Then, we predict the metric to incorporate with classification and centerness scores for better perceiving depth in the NMS procedure. And more importantly, we find that the prediction depth quality score can adaptively perceive outliers of depth prediction. It is the key for us to formulate a model-aware easy depth sample mining strategy, which alleviates the ill-posed issue of depth learning in monocular scenes. It brings out the potential of sample mining techniques.

\section{METHODOLOGY}
\label{method}

In this section, we briefly analyze several existing sample mining methods and apply them to depth mining. In sec.~\ref{method_MPM} and sec.~\ref{method_GAM}, we provide a detailed introduction to our mining approach. Before that, we first examine depth mining, as presented in subsection sec.~\ref{method_ana}.

\subsection{Analysis on Depth Mining}
\label{method_ana}

We mainly focus on easy mining and hard mining manners to explore the effect of depth mining on monocular 3D detectors. The typical easy mining method~\cite{PISA,iou-balanced} is done by endowing higher quality samples with larger weight, while hard mining method~\cite{streamyolo1,streamyolo2} is the opposite. The overall optimal object can be formulated as follows Eq.~\ref{eq:2} and Eq.~\ref{eq:3}. 

\begin{equation}
    \label{eq:2}
    \hat{\omega}_{g} = \omega _g\cdot\frac{{\sum_{i = 1}^N {{L}^{task}_i} }}{{\sum_{i = 1}^N {{\omega_i}}{{L}^{task}_i}}}, (g=0,1, \cdots, N)
\end{equation}

\begin{equation}
    \label{eq:3}
    {{L}_{mining}} = \sum_{i = 1}^N {\hat{\omega}_{i}{L}^{task}_i},
\end{equation}
where ${L}^{task}_i$ can be instantiated to any task loss of object $i$, and ${\omega}_{i}$ is the reweight metric, which can be instantiated to the corresponding quality (e.g., IoU). For an easy mining framework, ${\omega}_{i}$ is positively correlated with prediction quality, while ${\omega}_{i}$ is negatively correlated with prediction quality in a hard mining way. The normalized form of Eq.~\ref{eq:2} is typically used to keep the total task loss the same as its non-mining counterpart. Finally, the normalized factor $\hat{\omega}_{i}$ acts as the weight of task loss ${L}^{task}_i$.

To apply the above two types of sample mining techniques to depth learning, we instantiate ${\omega}$ as $\mathrm {DQ}^{+}$  and $\mathrm {DQ}^{-}$ for easy and hard mining, respectively. They derive from our defined depth quality metric as:

\begin{equation}
    \label{eq:4}
    \mathrm{DQ}^{+} = \mathrm{DQ}, \mathrm{DQ}^{-} = \frac{1}{\mathrm{DQ}} - 1,
\end{equation}
where $\mathrm {DQ}^{+}$ and $\mathrm {DQ}^{-}$ are substitute into Eq.~\ref{eq:2} and Eq.~\ref{eq:3} for carry out easy and hard depth mining supervision.

We conduct several experiments to reveal the effectiveness of depth sample mining. As shown in Tab.~\ref{table: fcos3d_reweight}, all experiments are based on FCOS3D~\cite{fcos3d} framework. The results demonstrate that an easy mining method can boost slight performance but not achieve qualitative change. As for the hard mining one, it significantly damages the detection performance. It triggers a big confusion: \emph{why sample mining works better in 2D dense prediction detectors for location regression but behaves badly in 3D dense prediction detectors for depth learning?} Our preliminary hypothesis is that this is caused by \emph{outliers} of prediction depth. Measuring the quality of depth is artificially subjective, but it is actually an ill-posed issue for the monocular framework to perceive depth quality. Therefore, rough increasing supervision of subjective easy samples may aggravate the difficulty of optimization. This result motivates us to alleviate the influence of outliers for the easy depth sample mining procedure.

\begin{table} 
\centering
\caption{Different depth sample mining methods. We use FCOS3D as the baseline detector.}
\begin{tabular}{l|c|c}
\toprule
Method & mAP & NDS \\
\midrule
Baseline~\cite{fcos3d} & 0.268 & 0.351 \\
\midrule
Subjective Mining ~\cite{PISA} & 0.270 & 0.356 \\
Relative improvement & +0.2\% & +0.5\% \\
\midrule
Hard Mining~\cite{streamyolo1} & 0.255 & 0.339\\
Relative improvement & -1.3\% & -1.2\% \\
\midrule
Model Perceive Mining (Ours) & 0.278 & 0.362 \\
Relative improvement & +1.0\% & +1.1\% \\
\midrule
GMM (Ours) & \textbf{0.286} & \textbf{0.370}\\
Relative improvement & \textbf{+1.8\%} & \textbf{1.9\%} \\
\bottomrule
\end{tabular}

\label{table: fcos3d_reweight}
\end{table}

\subsection{Preliminary Definition: Depth Quality Metric}

\begin{figure}[t]
\centering
\includegraphics[width=1.0\columnwidth]{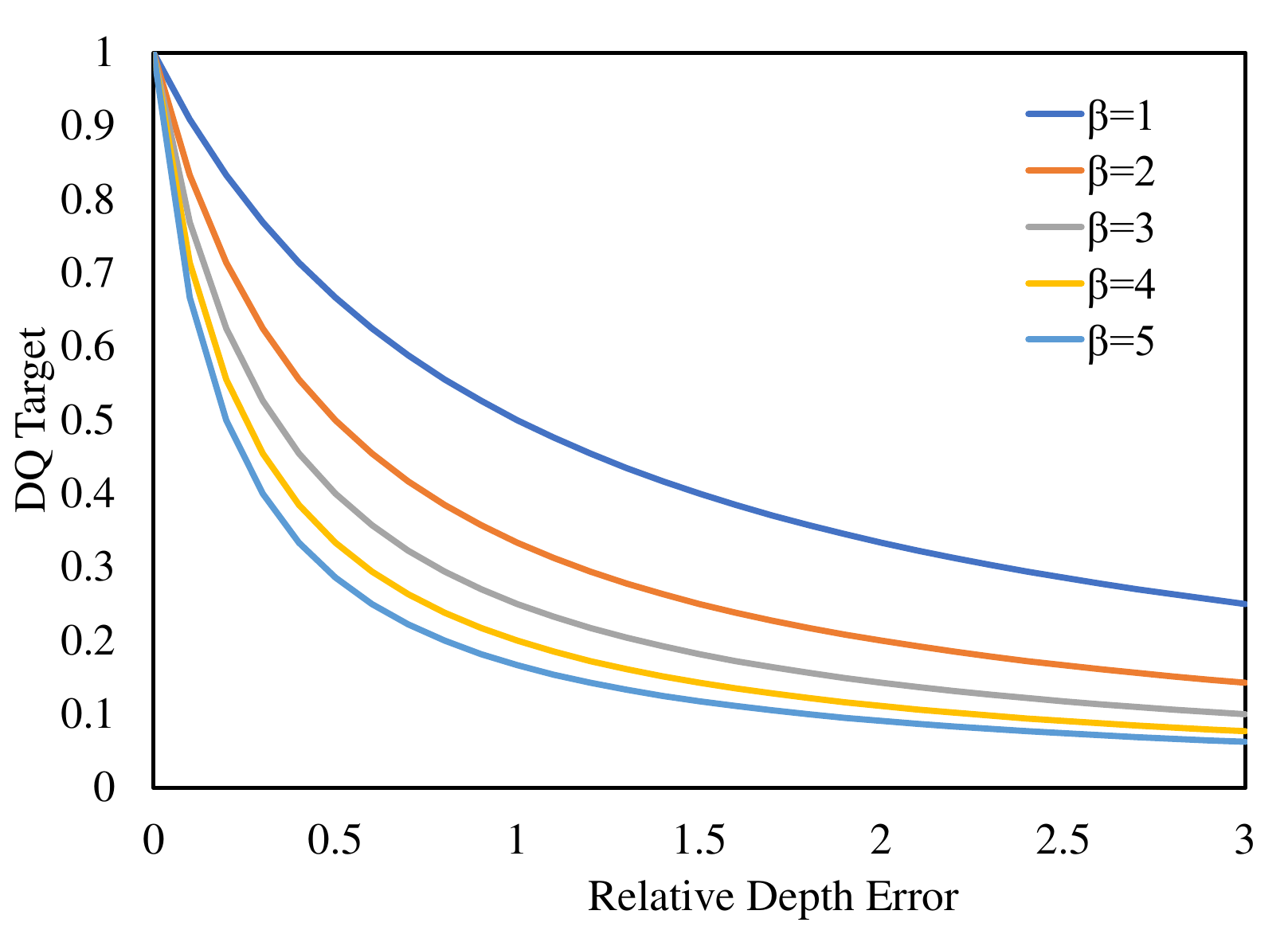}
\vspace{-.3in}
\caption{Variation in depth quality for different hyperparameters $\beta$. The smaller the $\beta$ value, the more sensitive the depth quality.}
\label{fig:DQ_target}
\end{figure}

As mentioned in Sec.~\ref{Introduction}, most existing methods employ prediction quality as a metric for the sample mining process. Along this line, we also design a metric to measure the depth quality. To do this, it needs to map the prediction error to normalized space, i.e., [0, 1]. Given prediction depth $\mathrm D_p$ and ground truth depth $\mathrm D_g$, we formulate depth quality metric $\hat{\mathrm{DQ}}$ as: 

\begin{equation}
\begin{aligned}
\hat{\mathrm{DQ}} = \frac{1}{\beta \frac{|\mathrm {D}_p - \mathrm {D}_g|}{\mathrm {D}_g} + 1}
\label{eq: 1}
\end{aligned}
\end{equation}

where $\hat{\mathrm{DQ}}$ becomes larger when the prediction depth error is smaller. $\beta$ is a hyper-parameter that controls the quality score distribution for depth estimation error. We set the difference between the ground truth depth and the predicted depth as the depth error $|\mathrm{D}_p - \mathrm{D}_g|$, and the relative depth error as $\frac{|\mathrm {D}_p - \mathrm {D}_g|}{\mathrm {D}_g}$. The distributions of the different quality scores regarding the relative depth error are shown in Fig.~\ref{fig:DQ_target}. It indicates that the quality value changes with greater sensitivity as $\beta$ decreases. We can also refer to this depth quality metric as a relative depth quality metric. In fact, the depth quality metric is not unique. we have also designed a Gaussian depth quality metric, the formula of which is expressed as follows:

\begin{equation}
\begin{aligned}
\hat{\mathrm{DQ}} = exp({- \frac{(D_{g}-D_{p})^2} {2{\beta}^2}})
\label{eq: gaussian_score}
\end{aligned}
\end{equation}

Where the meaning of $D_g, D_p, \beta$ is the same as in the relative depth quality metric mentioned earlier. Both metrics demonstrate good performance, with the relative depth quality metric being more suitable for PGD~\cite{pgd}, and the Gaussian depth quality metric being more suitable for FCOS3D~\cite{fcos3d}.

\subsection{Model Perceive Mining (MPM)}
\label{method_MPM}

In this section, we aim to endow the model with the ability to perceive depth's outliers. The outliers in this paper are defined as which depth's predictions are hard ill-posed objects. To adaptively capture this outlier, we extra propose predicting the depth quality with the above-defined DQ metric. These estimated bad depth qualities can reflect the existence of outliers to a certain extent. Technically, we append a branch head parallel to the depth branch to learn DQ. As shown in Fig.~\ref{fig:framework}(c), we use binary cross-entropy loss for the depth quality learning:

\begin{equation}
    \begin{aligned}
{L}_{dq} &= -\frac{1}{N}\sum_{i=1}^{N}[{{\hat{\mathrm{DQ_i}} \cdot \log{\mathrm{DQ_i}}}} \\
&+ {(1 - \mathrm{DQ_i}) \cdot \log{(1 - \hat{\mathrm{DQ_i}})}}],
    \end{aligned}
\label{Eq5}
\end{equation}
where $\hat{\mathrm{DQ}}$ is the ground truth quality calculated by Eq.~\ref{eq: 1}. N is the total number of positive samples.

As mentioned above, the prediction depth quality can perceive outlier, so it is natural to act as the reweight factor for easy mining learning. To this end, we instantiate $\omega$ of Eq.~\ref{eq:2} to our predicted DQ, smooth L1 loss for depth prediction task loss, and conduct Eq.~\ref{eq:3}. As shown in Fig.~\ref{fig:framework}(c), the overall depth learning includes BCE loss and reweight loss for depth supervision and depth sample mining supervision, respectively. The result of MPM is reported in Tab.~\ref{table: fcos3d_reweight}, it further boost the performance up to 1\% mAP and 1.1\% NDS. In contrast, Subjective Mining is virtually ineffective, because it relies entirely on depth prediction, which is inherently ill-posed.


\begin{figure}[t]
\centering
\includegraphics[width=1.0\columnwidth]{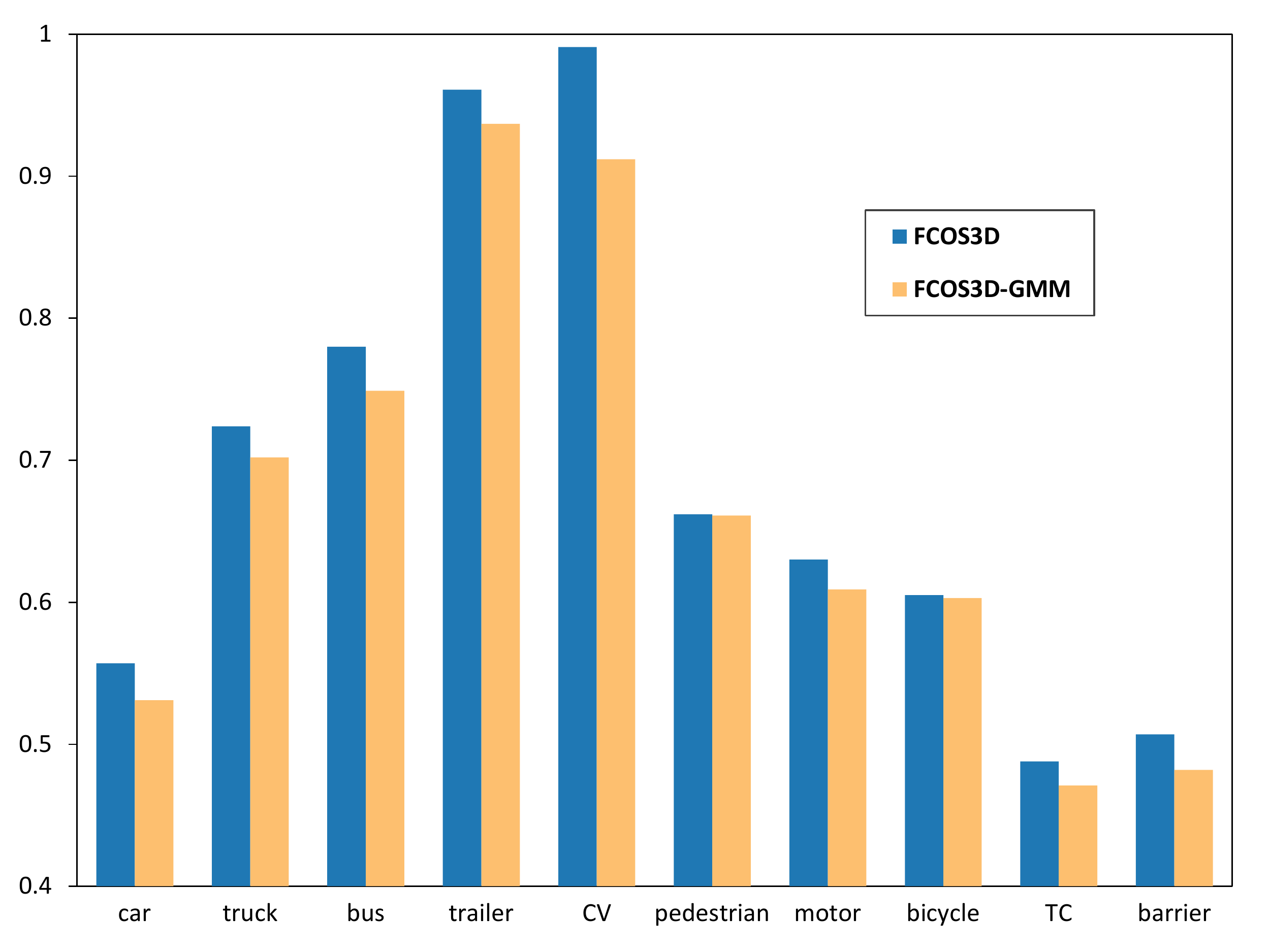}
\vspace{-.2in}
\caption{ Our GMM method reduces the ATE for each class, especially for large objects like construction vehicles (CV). Average Translation Error (ATE) is the Euclidean center distance in 2D. The decrease in mATE represents an improvement in the detector's depth estimation.}
\label{fig:mATE}
\end{figure}

\subsection{Gradient Aware Depth Mining (GAM)}
\label{method_GAM}

Going one step further, we introduce GAM for carrying out outlier-aware DQ prediction and easy implicit depth sample mining in a unified framework. As shown in Fig.~\ref{fig:framework}(d), it only needs to \emph{enable the gradient backward process} for DQ ground truth in BCE loss. To reveal the essence of GAM, we derive the gradient Eq.~\ref{eq:6} of the BCE loss (Eq.~\ref{Eq5}) with respect to $\hat{\mathrm{DQ}}$. 

\begin{equation}
	\begin{aligned}
\frac{\partial L_{BCE}}{\partial \hat{\mathrm {DQ}}} = \log (\frac{1-\mathrm {DQ}}{\mathrm {DQ}}).
	\end{aligned}
\label{eq:6}
\end{equation}

When DQ is larger than 0.5, it means samples most fall into the category of positive ones. When prediction depth quality increases from 0.5 to 1, the gradient magnitude increases continuously. It carries out an easy mining procedure, which guides the model to focus more on high-quality depth samples. Fig.~\ref{fig:DQ_target} shows, the effect of different values of $\beta$, on the curves of DQ target and relative depth error.


\subsection{Bonus: Depth Aware NMS} 

Existing works~\cite{fcos3d, pgd} only adopt the centerness score to alleviate the imbalance between classification score and prediction accuracy of the 3D box. We argue that it will make the 3D detector fall into sub-optimal performance since centerness score lacks depth information. To tackle this dilemma, we explore incorporating our predicting DQ with origin classification score and centerness score, guiding the NMS procedure to consider more depth clues for removing duplicate 3D detection boxes. Inspired by~\cite{iounet,iouaware,scd}, the NMS metric is reformulate as:

\begin{equation}
\begin{aligned}
s = \sqrt{s_{cls} \cdot s_{ctr} \cdot s_{DQ}},
\label{eq: final_scores}
\end{aligned}
\end{equation} 
where $s_{cls}, s_{ctr}, s_{DQ}$ denote as class scores, centerness and depth quality, respectively. Different from~\cite{iouaware}, we treat three of them as equally important rather than carrying out heavy effort to adjust hyper-parameters.

\begin{table}[t] 
\centering
\caption{Ablation study for the depth-aware scoring with FCOS3D on nuScenes validation.}
\begin{tabular}{c c c|c | c}
\toprule
Centerness & DQ & GAM & mAP & NDS \\
\midrule
 & & & 0.258 & 0.342 \\
\checkmark & &  & 0.268 & 0.351\\
 & \checkmark & & 0.272 & 0.354\\
& \checkmark & \checkmark &  0.276 & 0.361 \\
\checkmark & \checkmark & \checkmark & \textbf{0.286} & \textbf{0.370}\\
\bottomrule
\end{tabular}
\label{table: fcos3d_ds}
\end{table}

\begin{table}[t] 
\caption{Ablation study for the depth-aware scoring with different methods on nuScenes validation.}
\centering
\begin{tabular}{c |c | c | c}
\toprule
Method & mAP & NDS & mATE\\
\midrule
FCOS3D & 0.268 & 0.351 & 0.817\\
 (+GMM) & \textbf{0.286} & \textbf{0.370} & \textbf{0.779}\\
 \midrule
PGD & 0.293 & 0.373 & 0.768 \\
 (+GMM) & \textbf{0.303} & \textbf{0.379} & \textbf{0.757}\\
\bottomrule
\end{tabular}
\label{table: diff_methods}
\end{table}

\begin{table*} [htbp] \small
\centering
\caption{\textbf{Comparison of different paradigms on the nuScenes val set.} The results of FCOS3D and PGD are fine-tuned and tested with test time augmentation. $\dag$ indicates trained for 12 epochs (1x schedule) and ResNet-101 backbone. The BEVDet-R101 and BEVDet-STTiny are trained with CBGS~\cite{cbgs}. }
\begin{tabular}{l|c|c|ccccccc}
\toprule
Methods & Split & Modality & mAP$\uparrow$ & mATE$\downarrow$ & mASE$\downarrow$ & mAOE$\downarrow$ & mAVE$\downarrow$ & mAAE$\downarrow$ & NDS$\uparrow$\\
\midrule
\midrule
PointPillars(Light) & test & LiDAR & 0.305 & 0.517 & 0.290 & 0.500 & 0.316 & 0.368 & 0.453 \\
CenterFusion & test & Cam. \& Radar & 0.326 & 0.631 & 0.261 & 0.516 & 0.614 & 0.115 & 0.449 \\
CenterPoint v2 & test & Cam. \& LiDAR \& Radar & 0.671 & 0.249 & 0.236 & 0.350 & 0.250 & 0.136 & 0.714 \\
\midrule
CenterNet & val & Monocular & 0.306 & 0.716 & 0.264 & 0.609 & 1.426 & 0.658 & 0.328 \\
BEVDet-R101 & val & Multi-view & 0.317 & 0.704 & 0.273 & 0.531 & 0.940 & 0.250 & 0.389 \\
BEVDet-STTiny & val & Multi-view & 0.349 & 0.637 & 0.269 & 0.490 & 0.914 & 0.268 & 0.417 \\ 
FCOS3D & val & Monocular & 0.343 & 0.725 & 0.263 & 0.422 & 1.292 & 0.153 & 0.415\\
PGD & val & Monocular & 0.369 & 0.683 & 0.260 & 0.439 & 1.268 & 0.185 & 0.428\\
\midrule
LRM0 & test & Monocular & 0.294 & 0.752 & 0.265 & 0.603 & 1.582 & 0.14 & 0.371 \\
MonoDIS & test & Monocular & 0.304 & 0.738 & 0.263 & 0.546 & 1.553 & 0.134 & 0.384 \\
CenterNet & test & Monocular & 0.338 & 0.658 & 0.255 & 0.629 & 1.629 & 0.142 & 0.400 \\
Noah CV Lab & test & Monocular & 0.331 & 0.660 & 0.262 & \textbf{0.354} & 1.663 & 0.198 & 0.418 \\
FCOS3D & test & Monocular & 0.358 & 0.690 & 0.249 & 0.452 & 1.434 & 0.124 & 0.428 \\
PGD & test & Monocular & 0.386 & 0.626 & 0.245 & 0.451 & 1.509 & 0.127 & 0.448 \\
\midrule
GMM (Ours) & test & Monocular & \textbf{0.421} & \textbf{0.614} & \textbf{0.243}	& 0.395 & \textbf{1.171} & \textbf{0.126} & \textbf{0.473}\\

\bottomrule
\end{tabular}
\label{table: nuscenes_results}
\end{table*}

\begin{table*}[ht] \small 
\centering
\caption{\textbf{Ablation studies on the nuScenes 3D detection benchmark.} In each method, the results without $\checkmark$ are those of FCOS3D, while with $\checkmark$ are those of adding our GMM method. The results show the robustness of our method to different backbone networks and techniques. }
\begin{tabular}{c|c|ccccccc}
\toprule
Methods & GMM & mAP$\uparrow$ & mATE$\downarrow$ & mASE$\downarrow$ & mAOE$\downarrow$ & mAVE$\downarrow$ & mAAE$\downarrow$ & NDS$\uparrow$ \\
\midrule
\midrule
\multirow{2}*{Baseline (w/ ResNet-50)} & & 0.268 & 0.817 & 0.271 & 0.586 & 1.315 & 0.156 & 0.351\\
& \checkmark & 0.286 &	0.779 &	0.264 & 0.540 & 1.319 & 0.153 & 0.370 \\
\midrule
\multirow{2}*{+ Stronger backbone (ResNet-101)} & & 0.280 & 0.822 & 0.274 & 0.640 & 1.305 & 0.177 & 0.349 \\
& \checkmark & 0.308 & 0.778 & 0.265 & 0.516 & 1.184 & 0.167 & 0.382 \\
\midrule
\multirow{2}*{+ DCN in backbone} & & 0.295 & 0.806 & 0.268 & 0.511 & 1.315 & 0.170 & 0.372 \\ 
& \checkmark & 0.309 & 0.754 & 0.266 & 0.510 & 1.244 & 0.169 & 0.385 \\
\midrule
\multirow{2}*{+ Finetune w/ depth weight=1.0} & & 0.316 & 0.755 & 0.263 & 0.458 & 1.307 & 0.169 & 0.393 \\
& \checkmark & 0.327 & 0.733 & 0.261 & 0.489 & 1.179 & 0.160 & 0.399\\
\midrule
\multirow{2}*{+ Test time augmentation} & & 0.326 & 0.743 & 0.259 & 0.441 & 1.341 & 0.163 & 0.402\\
& \checkmark & 0.340 & 0.722 & 0.260 & 0.499 & 1.156 & 0.160 & 0.406\\
\midrule
\multirow{2}*{+ More epochs \& ensemble} & & 0.343 & 0.725 & 0.263 & 0.422 & 1.292 & 0.153 & 0.415 \\
& \checkmark & 0.348 & 0.711  & 0.262 & 0.455 & 0.984 & 0.161  & 0.417\\
\midrule
\multirow{2}*{\textbf{Test result}} & & 0.358 & 0.690 & 0.249 & 0.452 & 1.434 & 0.124 & 0.428 \\
& \checkmark & \textbf{0.376} & \textbf{0.666} &\textbf{ 0.242} & 0.464 & 1.149 & \textbf{0.121} & \textbf{0.439}\\
\bottomrule
\end{tabular}
\label{table: fcos3d_ablation}
\end{table*}

\section{EXPERIMENTS}
\subsection{Datasets and Metrics}
\noindent\textbf{Nuscenes Datasets} We evaluate our approach using nuScenes~\cite{nuscenes}, a large-scale, widely used dataset. It is made up of multi-modal data gathered from 1000 scenes, including RGB pictures from six surround-view cameras, points from 5 radars, and one LiDAR. For training/validation/testing, it is divided into 700/150/150 scenes. There are 1.4 million annotated 3D bounding boxes from ten categories in total. It is quickly becoming one of the most authoritative benchmarks for 3D object detection due to its variety of scenes and ground truths. As a result, we use it to validate the efficacy of our technique.

\noindent\textbf{Metrics:} The official metrics for the Nuscenes dataset are mAP (mean Average Precision) and NDS. Below, we briefly describe these two metrics.
The most frequent measure for object detection is mAP with an IoU threshold. 3D object detection scenes and tasks are more complex, mAP can not capture all aspects detection tasks, like rotation, velocity. Therefore, Nuscenes propose a different score: the nuScenes detection score (NDS). 

\begin{equation}
NDS = \frac{1}{10} [5mAP + \sum_{mTP\in \mathbb{TP}} (1 - \min(1,mTP))],
\label{eq: NDS}
\end{equation}
where TP is the set composed of five True Positive metrics. mTP is consist of mean Average Translation Error(mATE), mean Average Scale Error(mASE), mean Average Orientation Error(mAOE), mean Average Velocity Error(mAVE), 
mean Average Attribute Error(mAAE).Since mAVE, mAOE and mATE can be larger than 1, Nuscenes bound each metric between 0 and 1.

\subsection{Implementation Details}
We use our reproduced FCOS3D~\cite{fcos3d} and PGD~\cite{pgd} for all experiments. For all ablation experiments, unless specifically stated, we employ ResNet-50~\cite{resnet} based Feature Pyramid Networks (FPN)~\cite{fpn} as the feature extraction backbone for generating multi-level predictions. For the input, input images are resized into 1600px $\times$ 900px for both training and testing. In image-based detectors, shared detection heads between FPN levels often achieve greater performance, and we follow this setup.  All convolutional modules consist of basic convolution, batch normalization, and activation layers, with normal distribution used for weight initialization. The overall framework is built on top of MMDetection3D~\cite{mmdet3d2020}. Following the approach in FCOS3D, we utilize a weight of 0.2 for depth regression during training to enhance stability. For more competitive performance and accurate detection, we fine-tune our model with this weight set to 1.

\noindent\textbf{Training Parameters.} For all the experiments, we trained randomly initialized networks from scratch following end-to-end manners. Models are trained with SGD optimizer, employing gradient clipping and a warm-up policy with a learning rate of 0.002, 500 warm-up iterations, a warm-up ratio of 0.33, and a batch size of 16 on 8 GTX 2080Ti GPUs for nuScenes~\cite{nuscenes}. Besides, during fine-tuning, we adjust the learning rate from 0.002 to 0.001 and change the default depth weight from 0.2 to 1. Regarding data augmentation, we follow the default settings of PGD and FCOS3D. Related results are presented in the ablation study.

For online testing submission, we adopt two widely-used settings: test-time augmentation (TTA) and model ensemble. Specifically, for the PGD-GMM online submission model, we employ ConvNext-Base~\cite{ConvNeXt22} as the backbone for image feature extraction. This backbone is initialized from the instance segmentation model Cascade Mask R-CNN~\cite{CascadeRCNN} pretrained on nuImage~\cite{nuscenes}, following the Transfusion~\cite{TransFusion2022}.

\begin{figure}[t]
\centering
\includegraphics[width=1.0\columnwidth]{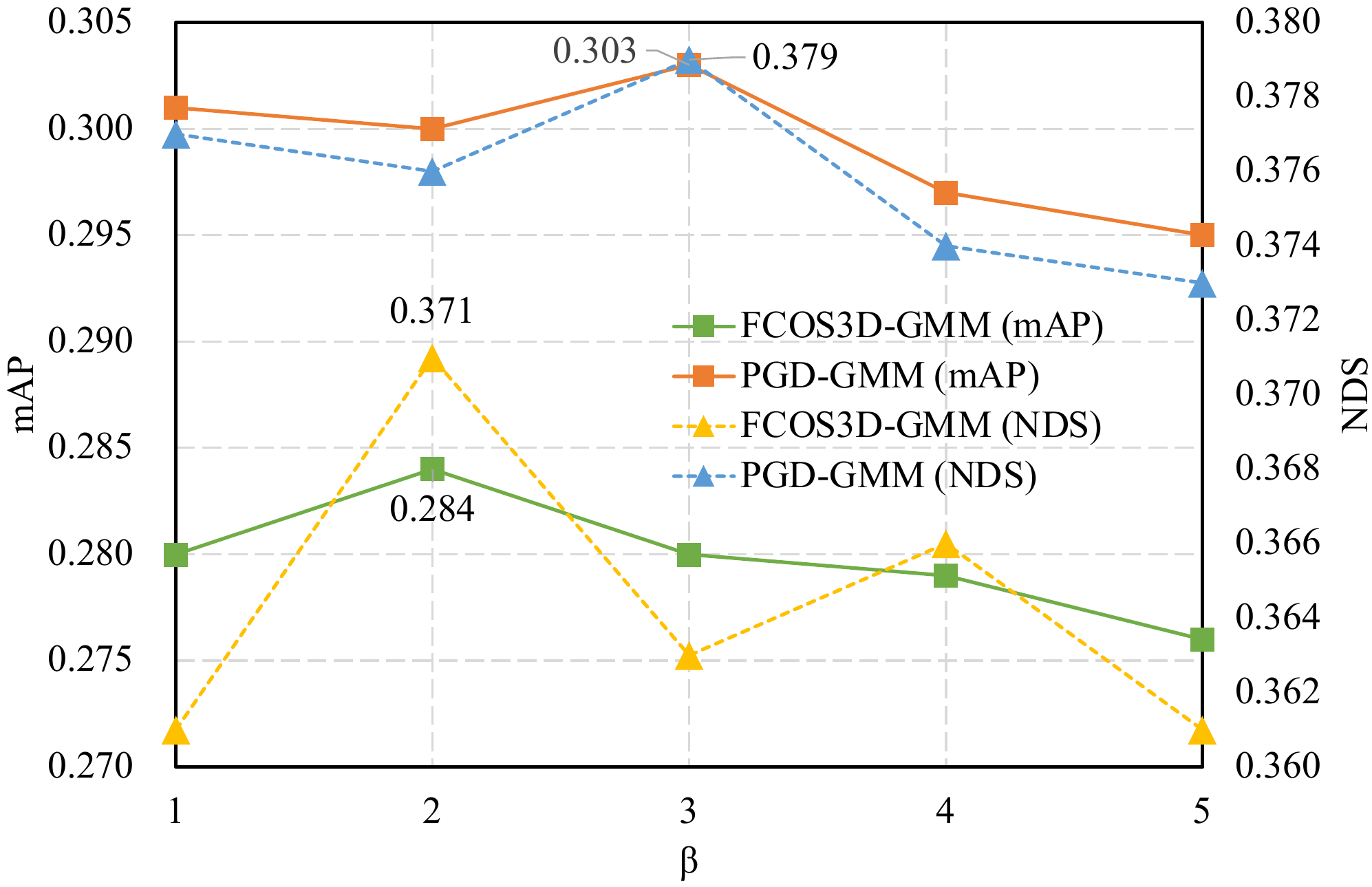}
\vspace{-.2in}
\caption{Ablating hyper-parameter $\beta$ (in Eq.~\ref{eq: 1}) for depth target. We report the mAP and NDS metrics for FCOS3D and PGD 3D detectors.}
\label{fig:beta}
\end{figure}

\subsection{Comparison with State-of-the-Arts}
Initially, we present the results of the quantitative analysis on the nuScenes~\cite{nuscenes} dataset in Table~\ref{table: nuscenes_results}. It is evident that our method (GMM) surpasses the previous state-of-the-art monocular methods, such as PGD. Our method achieves a impressive performance of 42.1\% mAP and 47.3\% NDS. We also observe substantial improvements in mAVE and mAOE. As shown in the test results of Table~\ref{table: fcos3d_ablation}, under the exact same settings, our FCOS3D-GMM significantly outperforms FCOS3D, achieving 37.6\% mAP and 43.9\% NDS.

On the validation set, we first compared all of the approaches by using RGB photographs as input data. 
It is clear that our method has significantly improved both the original FCOS3D and PGD. In particular, FCOS3D-GMM achieves 34.0\% mAP and 40.6\% NDS, which beats FCOS3D by 1.4\% in terms of mAP in the 5th row of Table~\ref{table: fcos3d_ablation}. Indeed, our approach has huge advantages over LRM0, MonoDis~\cite{monoDis}, Noah CV Lab and centreNet~\cite{centerpoint}. Additionally, we only use the single-frame image in our experiments while BEVDet~\cite{bevdet} uses six frames from 6 different cameras as input. 
Due to computational resource constraints, our experiments are conducted using a 1x training schedule. On the other hand, BEVDet uses the CBGS augmentation strategy, and this can lead to huge gains. 


\subsection{Ablation Study}
In this section, we perform ablations on some important components base on nuScenes dataset.

\paragraph{Depth-aware Scoring} Table~\ref{table: fcos3d_ds} presents the effects of depth-aware scoring. It is evident that centerness leads to a gain of 1\% and 0.9\% on mAP and  NDS, respectively, while our depth-aware quality (DQ) results in a gain of 1.4\% and 1.2\% in mAP and NDS compared to the baseline without centerness. Since Gradient Aware Depth Mining (GAM) requires a depth quality prediction branch, we typically use DQ and GAM together, which achieves an mAP of 27.6\%, NDS of 36.1\%. Most importantly, we combine the above, and the model attains 28.6\% mAP and 37\% NDS, suggesting that centerness prediction and DQ prediction do not conflict.

\paragraph{Depth-aware gradient mining} Table~\ref{table: fcos3d_reweight} depicts the impact of various gradient mining approaches. Subjective Mining illustrates that methods such as~\cite{PISA, GHM, iou-balanced} do not significantly improve model performance. Furthermore, we compare model perceive Mining with baseline, and our method has a 1\% improvement on mAP and a 1.1\% improvement on NDS. Finally, Union Model Perceive Mining, which combines depth-aware scores and Model Perceive Mining, which has remarkable improvements of 1.8\% for mAP and 1.9\% for NDS compared to baseline. In particular, we can see a significant improvement in mATE, with a 4.8\% drop from 81.7\% to 76.9\%. Average Translation Error (ATE) is the Euclidean center distance in 2D, which is strongly correlated with depth. Therefore, the considerable decrease in ATE confirms the credibility of our motivations. 

\paragraph{Other critical factors} We report our results on different backbone networks, including ResNet-50/101/101+DCN~\cite{dcn}, as shown in Table~\ref{table: fcos3d_ablation}. It can be observed that even with a stronger backbone, our method continues to perform exceptionally well within the FCOS3D framework. Additionally, we employ fine-tuning and Test Time Augmentation (TTA), resulting in 34\% mAP and 40.6\% NDS. It is worth noting that more epochs refer to a 2x training schedule, and the model we submit to the nuScenes~\cite{nuscenes} test online board uses the same strategy. Table~\ref{table: diff_methods} demonstrates that our method remains effective across various model architectures (FCOS3D and PGD).
Furthermore, We design the depth quality target with a hyperparameter $\beta$. We experiment with different parameter values and from Figure~\ref{fig:beta}, it is determined that using $\beta = 2$ in FCOS3D and $\beta = 3$ in PGD yields the most suitable results. \label{fig:mATE} shows the enhancement of GMM for different categories of ATE.

\section{CONCLUSIONS}
In this paper, we conduct extensive experiments to explore the effectiveness and limitations of existing sample mining techniques for depth perception of 3D monocular detectors during training. It reveals that the bottleneck towards realizing its potential is the failure to tackle outliers in prediction depth. To address this issue, we introduce a Model Perceive Mining (MPM) method to adaptively excavate outliers and reduce its contribution during training, enabling us to transform the ill-posed problem for depth learning into a soft ill-posed one. Going a step further, we introduce a novel Gradient-Aware Depth Mining (GAM) strategy to replace traditional mining manner, and it simultaneously generalizes the MPM procedure. Extensive experiments demonstrate the efficacy of our method, which achieves state-of-the-art performance in 3D monocular detection tasks. Extensive experiments demonstrate the efficacy of our method, which achieves state-of-the-art performance in 3D monocular detection tasks. We believe our work can inspire more researchers to focus on sample mining techniques for 3D detection tasks.

\ifCLASSOPTIONcaptionsoff
  \newpage
\fi

{
\bibliographystyle{ieeetr}
\bibliography{sample.bib}
}

\end{document}